\documentclass{article}

    \usepackage[final,nonatbib]{neurips_2020}

\usepackage[utf8]{inputenc} % allow utf-8 input
\usepackage[T1]{fontenc}    % use 8-bit T1 fonts
\usepackage[hyperfootnotes=false]{hyperref}       % hyperlinks
\usepackage{url}            % simple URL typesetting
\usepackage{booktabs}       % professional-quality tables
\usepackage{amsfonts}       % blackboard math symbols
\usepackage{nicefrac}       % compact symbols for 1/2, etc.
\usepackage{microtype}      % microtypography
\usepackage{amsmath}
\usepackage{graphicx}
\usepackage{array,multirow}
\usepackage{color}
\usepackage{url}
\newcommand{\first}[1]{\textcolor{red}{{\textbf{#1}}}}
\newcommand{\second}[1]{\textcolor{blue}{{\underline{#1}}}}

\newcommand\blfootnote[1]{%
  \begingroup
  \renewcommand\thefootnote{}\footnote{#1}%
  \addtocounter{footnote}{-1}%
  \endgroup
}

\title{Video Frame Interpolation without Temporal Priors}

\author{%
	Youjian Zhang$^{*}$ \\
	The University of Sydney, Australia \\
	\texttt{\small yzha0535@uni.sydney.edu.au} \\
	\And
	Chaoyue Wang$^{*}$ \\
	The University of Sydney, Australia \\
	\texttt{\small chaoyue.wang@sydney.edu.au} \\
	\AND
	Dacheng Tao \\
	The University of Sydney, Australia \\
	\texttt{\small dacheng.tao@sydney.edu.au} \\
}

\begin{document}
	
	\maketitle
	
	\blfootnote{* indicates equal contribution.}
	
	\begin{abstract}

		Video frame interpolation, which aims to synthesize non-exist intermediate frames in a video sequence, is an important research topic in computer vision. Existing video frame interpolation methods have achieved remarkable results under specific assumptions, such as instant or known exposure time. However, in complicated real-world situations, the temporal priors of videos, \textit{i.e.,} frames per second (FPS) and frame exposure time, may vary from different camera sensors. When test videos are taken under different exposure settings from training ones, the interpolated frames will suffer significant misalignment problems. In this work, we solve the video frame interpolation problem in a general situation, where input frames can be acquired under uncertain exposure (and interval) time. Unlike previous methods that can only be applied to a specific temporal prior, we derive a general curvilinear motion trajectory formula from four consecutive sharp frames or two consecutive blurry frames without temporal priors. Moreover, utilizing constraints within adjacent motion trajectories, we devise a novel optical flow refinement strategy for better interpolation results. Finally, experiments demonstrate that one well-trained model is enough for synthesizing high-quality slow-motion videos under complicated real-world situations. Codes are available on \url{https://github.com/yjzhang96/UTI-VFI}.
	\end{abstract}
	
	\section{Introduction}\label{Sec:Intro}
	Video frame interpolation aims to synthesize non-exist intermediate frames and thereby provides a visually fluid video sequence. It has broad application prospects, such as slow motion production~\cite{jin2019learning}, up-converting frame rate~\cite{castagno1996method} and novel-view rendering~\cite{flynn2016deepstereo}.
	
	Many state-of-the-art video interpolation methods \cite{bao2019depth,jiang2018super,liu2017video,xu2019quadratic} aim to estimate the object motion and occlusion with the assistance of optical flow. Through refining forward and backward motion flows among several frames, these methods can directly warp pixels to synthesize desired intermediate frames. %Therefore, exploring even more accurate dense motion flow becomes one of the fundamental problems of VFI related tasks. 
	To achieve this goal, some popular datasets, of which either triplet images or 240fps high-frame-rate videos, are collected as the ground-truth of real-world motions. Meanwhile, to evaluate the performance of proposed methods, the well-trained model is tested using frames collected in a similar way. Although significant improvement has demonstrated by experiments of recent works, people may ask if the same (or similar) performance can be achieved in complicated real-world situations.

	%Among them, \cite{jiang2018super,xu2019quadratic} are vary practical for increasing the frame rate by arbitrary times, since they can easily calculate the position of the object given the time step and two adjacent reference frames. 
	
	%However, when we try to convert a low-frame-rate video to a high-frame-rate one, these methods have a problem accurately estimating the optical flow since the frames in the low-frame-rate video are usually blurred. This divergence between real scenarios and training data caused by exposure time leads to a significant deterioration in former video interpolation methods. To solve this problem, some methods \cite{jin2019learning,shen2020blurry} are proposed to conduct deblurring and frame interpolation jointly. However, both of the methods cannot be applied generally to arbitrary filming parameter setting as we will discuss in the next paragraph, and can only produce a promising result in their own setup.
	
	%Most interpolation methods are trained and tested on some popular datasets, which are either triplets images or 240fps high-frame-rate videos. When the frames in these datasets are treated as source video, there underlies an assumption that these video frames are sharp, which means the exposure time is instant.(This dose not conform with the setup of low-frame-rate video which has a relatively long exposure time.)
	
\begin{figure}[t!]
		\centering
		\includegraphics[width=\linewidth]{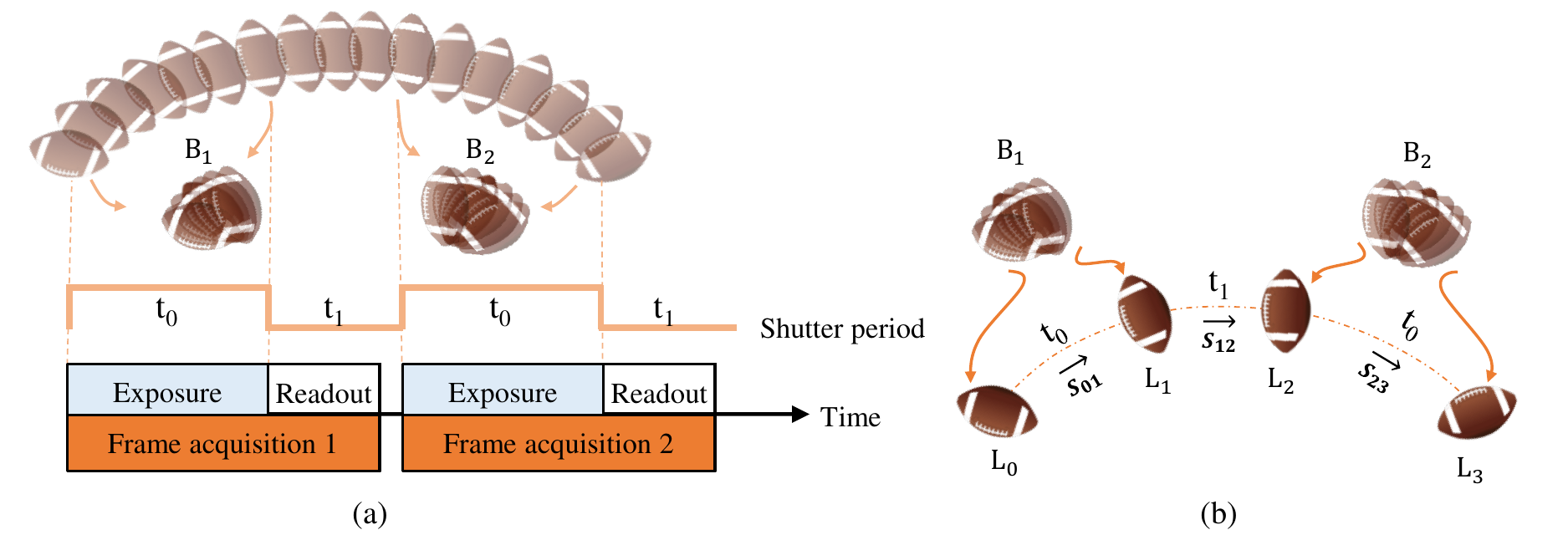}
		\caption{(a) Illustration of frame acquisition in video shooting. In real-world situation, the time interval $t_0$ and $t_1$ are unknown and may vary under different exposure setting. In the specific example in the figure, the time intervals are set to $t_0/t_1=6:4$, which indicates the intra-frame and inter-frame interpolation should be 7 and 3 frames respectively when we want to interpolate 10 frames. (b) Our proposed method aims to determine the uncertain time intervals and perform interpolation from the four consecutive states.} \vskip -0.2 cm
		\label{fig:exposure}
	\end{figure}

	%For a better interpretation of our research problem
	To comprehensively discuss this question, we first revisit the principle of video frame acquisition. As illustrated in Fig.~\ref{fig:exposure}~(a), the frame acquisition process usually includes two phases: exposure phase and readout phase. In the exposure phase, the shutter opens for a duration of $ t_{0} $ so that the photosensitive sensor is exposed. In the readout phase, the camera reads the charge on the pixel array and convert the signal to get the pixel value. Considering different technologies of cameras, the readout phase could be either overlapped or non-overlapped with the exposure phase. Here, Fig. \ref{fig:exposure}~(a) is an example of non-overlapped exposure. For easy discussing, we define the time interval between two exposures as $t_{1}$. Thus, a complete shutter period is defined as the time period $ t_{0} + t_{1}$. Correspondingly, frames per second (FPS) is defined as the reciprocal of the shutter period.
	%Normally, FPS is restricted by the exposure time and readout time as the non-overlap mode shown in Fig. \ref{fig:exposure}. However, owning to the technology advance, the exposure phase can be overlapped with readout phase to increase frame rate, yet the time interval $ t_{1} $ cannot be eliminated because of the intrinsic demand of the sensor.
	Note that, $ t_{1} $ cannot be eliminated because of the intrinsic demand of the sensor. Meanwhile, $ t_{0} $ cannot be too short compared to shutter period, otherwise it will produce a visually discontinuous video.
	
	The exposure time $t_0$ and the interval time $t_1$ (or FPS $\frac{1}{t_0+t_1}$) are two important parameters of a camera sensor, and they could vary largely across different cameras~\cite{camerareadout}. Therefore, when we perform frame interpolation on real-world videos, following challenges should be further considered: 1) Due to the existence of exposure time, the movement of the camera and object may produce motion/dynamic blur within a video frame. Directly performing the interpolation between blurry frames would lead to inferior visual results. A more severe blur would usually occur in the lower frame-rate video since the exposure time is relatively long. 2) Simply combining deblurring and video interpolation techniques may not handle the blurry video frames well. For blurry video frames, we should not only focus on the inter-frame interpolation, but also perform the intra-frame interpolation. 3) Note that $ t_{0} $ and $ t_{1} $ may vary due to the limitation of equipment or different exposure settings, the number of interpolated frames and corresponding motion trajectories will vary accordingly. For example, in the instance of Fig~\ref{fig:exposure}~(a), if we want to up-convert the FPS by 10 times, we should interpolate 7 frames underlying each blurry frame, and 3 frames between the two consecutive frames. Similarly, the estimation of the motion trajectory must consider the uneven time intervals. According to our observation, most existing works cannot overcome these three challenges simultaneously. Although the most recent works \cite{jin2019learning, shen2020blurry} manage to solve the problem of motion blur in video interpolation, they are trained on the specific exposure setting and could be hard to generalize to different situations.
	
	%both cannot be applied to a more general situation because they neglect the second discussion as aforementioned. Specifically, they focus on a specific setting of $ t_{0} $ and $ t_{1} $, as a result, they will suffer a significant frame misalignment when the test video is filmed under different settings.
	
	To address these issues, in this work, we consider the video frame interpolation problem in a more general situation and aim to deliver more accurate interpolation results.
	%issues of generalized video frame interpolation, we propose a method which can overcome the uncertainty of $ t_{0} $ and $ t_{1} $, thus provide a more accurate prediction for motion trajectory.
	Specifically, giving a video sequence as the input, we first train a second-order residual key-states restoration network to synthesize the start and the end states for each frame, \textit{e.g.} $L_0$ and $L_1$ in Fig.~\ref{fig:exposure}~(b). If there exists zero movement (misalignment) between two states, the video frame is regarded as one instant frame (\textit{i.e.} without blur). Otherwise, the exposure time cannot be ignored, and both inter- and intra-interpolation are performed.
	%Contract to the former methods which only focus on the latent frame contents stored in the input blurry frames, we find that the blurry frames also encode the intrinsic exposure time and shutter period, \textit{i.e.} $ t_{0} $ and $ t_{1} $.
	%Specifically, we first restore two key-states which have invariant position regardless of the different exposure time setting. These key-stateses denote the start and end point of the blurry trajectory as shown in Fig. \ref{fig:exposure}(b).
	Moreover, following the same assumption as \cite{xu2019quadratic}, \textit{i.e.} the acceleration of motion remains consistent during consecutive frames, we apply the quadratic model~\cite{mcallister1978interpolation,xu2019quadratic} to the general video acquisition situation. We derive the general curvilinear motion representation without temporal priors from consecutive four key-states, such as $L_0$, $L_1$, $L_2$ and $L_3$ in Fig.~\ref{fig:exposure}~(b). Meanwhile, the relationship between $ t_{0} $ and $ t_{1} $ can be determined by the displacements between key-states, \textit{i.e.} $\boldsymbol{S_{01}}$, $\boldsymbol{S_{12}}$ and $\boldsymbol{S_{23}}$. In addition, to reduce the adverse effects caused by inferior optical-flow estimation,
	%With $ t_{0} $ and $ t_{1} $ acquired, we can determine the number of intermediate frames to be inserted between blurry frames or inside blurry frames.
	%Although the temporal information can be straightforward calculated through motion trajectories, the result is not satisfying since the inaccurate optical flow estimation will bring great errors to the estimation of time intervals and the following interpolation.
	%To address this problem, 
	we further refine the optical flows with the derived trajectory priors. 
	Finally, the refined optical flows are utilized to perform high-quality intermediate frame synthesis. 
	%Noted that when the video frames are sharp (instant exposure time), our system share the same solution with QVI \cite{xu2019quadratic}. %Thus, our system is designed for a more generalized video frame interpolation.
	
	Overall, in this paper, we make following contributions: 1) We propose a restoration network to synthesize start and end states of the input video frames. This network is able to handle different exposure settings, and remove blur in the original video clip; 2) We derive a curvilinear motion representation which is sensitive to different exposure settings, thereby providing a more accurate frame alignment for uncertain time interval interpolation; 3) We further refine the optical flow with the trajectory priors to improve the interpolation results. We construct different datasets to simulate the different exposure settings in real scenarios. Comprehensive experiments on these datasets and real-world vidoes demonstrate the effectiveness of our proposed framework.

	\section{Related Works}
	\textbf{Video frame interpolation.} Most popular video interpolation methods utilize optical flow \cite{jiang2018super,xu2019quadratic,liu2017video,bao2019depth,bao2019memc,xue2019video} to predict the motion for the interpolated frame. Some methods \cite{niklaus2017video,niklaus2017adap,gui2020featureflow} estimate space-varying and separable convolution filters for each pixel, and synthesize the interpolated pixel from a convolution between two adjacent patches. Xu \textit{et al.}~\cite{xu2019quadratic} proposes a quadratic interpolation to allow the interpolated motion to be curvilinear instead of being uniform and linear. However, all these methods will encounter difficulties when processing the blurry video since the optical flow/motion estimation will be inaccurate.
	%However, these data-driven methods are trained and tested on sharp video clips, which are not applicable for blurry videos since there is a huge divergence between sharp and blurry datasets. Also, the optical flow estimation will become inaccurate and further deteriorate the interpolation performance.
	
	\textbf{Video/Image deblurring.} Conventional video deblurring methods~\cite{fergus2006removing,hirsch2011fast,whyte2014deblurring} usually apply the deconvolution algorithm with the assistance of image priors or regulations. To make full use of adjacent frames, Hyun \textit{et al.} \cite{hyun2015generalized} utilize inter-frame optical flow to estimate blur kernels. Ren \textit{et al.} \cite{ren2017video} also apply optical flow to facilitate the segmentation result. More recently, deep convolutional neural networks (CNN) have been applied to bypass the restriction of blur type or image priors \cite{nah2017deep,su2017deep,gao2019dynamic,nah2019recurrent,kupyn2018deblurgan,zhang2018dynamic}, and enable an end-to-end training scheme by introducing the synthetic real-world scene datasets \cite{nah2017deep,su2017deep}. To exploit the temporal relationship, Nah \textit{et al.}~\cite{nah2019recurrent} propose a recurrent neural network (RNN) to iteratively update the hidden state for output frames. Wang \textit{et al.}~\cite{wang2019edvr} devise a pyramid, cascading and deformable alignment module to conduct a better frame alignment in feature level, and their method won the first place in the NITRE19 video deblurring challenge~\cite{nah2019ntire}. There are also some works~\cite{zhang2020selfsupervised,jin2018learning,purohit2019bringing} learning to extract a video clip from a blurry image, which can be considered as a combination of image deblurring with intra-frame interpolation.
	
	\textbf{Joint video deblurring and interpolation.} Recent methods~\cite{jin2019learning,shen2020blurry} have been proposed to address the blurry video interpolation problem. Among them, Jin \textit{et al.}~\cite{jin2019learning} first extract several keyframes, and then interpolate the middle frame from two adjacent frames. Meantime, Shen \textit{et al.}~\cite{shen2020blurry} proposed a joint interpolation method, where they simultaneously output the deblurred frame and interpolated frame in a pyramid framework. %However, they can only achieve an upsampling by 2 times. 
	Both these methods have pre-defined a specific setting for the blurry video exposure mechanism, which may fail when applied to a video acquired from other equipment or other camera settings.

	\section{The Proposed Video Interpolation Scheme}\label{Sec:UTI}
	%As we mentioned in Sec. \ref{Sec:Intro}, a generalized video interpolation method should tackle these two problems:(1) uncertain time intervals, (2) motion blur caused by exposure time. 
	To address the aforementioned challenges of video frame interpolation without temporal priors, in this section, we introduce the proposed interpolation scheme in detail. Firstly, to overcome the problem caused by the uncertainty of the time interval, we derive a new quadratic formula for different exposure settings. Then, utilizing the motion flow priors contained in the formula, we further refine the estimated optical flow for more accurate time interval and trajectory estimation. Finally, we introduce the second-order residual learning strategy for key-states restoration from input frame sequences.
	
	%we derived a novel formula to overcome the uncertainty of the time interval. 
	
	%and the priors for optical flow refinement. 
	%And we will discuss the approach to apply this formula in the blurry videos with different setting of exposure time in Sec \ref{Sec:3}.
	
	\subsection{From equal time interval to uncertain time interval}
	To interpolate intermediate frame $ L_{t} $ between two consecutive frames $ L_{1} $ and $ L_{2} $, the optical flow based video interpolation methods \cite{jiang2018super,liu2017video,xu2019quadratic} aim to estimate the optical flow from frame $ L_{1} $ to $ L_{t} $, or frame $ L_{2} $ to $ L_{t} $. %Contrast to former approaches which assume a linear motion between  $ L_{1} $  and $ L_{2} $ , 
	Recently, inspired by \cite{mcallister1978interpolation}, Xu \textit{et al.} \cite{xu2019quadratic} have relaxed the constrains of motion from linear displacement to quadratic curvilinear, which corresponds to acceleration-aware motions:
	\begin{equation}\label{eq:QVI}
	\boldsymbol{S_{1t}} = (\boldsymbol{S_{12}} -\boldsymbol{S_{01}})/2 \times t^{2} + (\boldsymbol{S_{12}} +\boldsymbol{S_{01}})/2 \times t,
	\end{equation}
	where $\boldsymbol{S_{ab}}$ means the displacement of pixels from frame $a$ to frame $b$, and it is calculated by optical flow. In order to keep the pixel coordinates aligned in each optical flow map, the start point of these optical flows should be the same. In general, the displacements are calculated as $ \boldsymbol{\hat{S}_{12}} = \boldsymbol{f_{1 \rightarrow 2}}, \boldsymbol{\hat{S}_{01}} = -\boldsymbol{f_{1 \rightarrow 0}}$, where $\boldsymbol{f_{a \rightarrow b}}$ denotes the optical flow from frame a to frame b.
	
	%However, in the derivation of this formula \ref{eq:QVI}, they assume the time intervals of every frames are equal, which is one unit time. Though this may not be a problem in most video interpolation tasks since most datasets have a frame sequences with uniform time intervals, we can still derive a more universal formula which relaxes the constrain of time. Here, we assume the time interval of 3 consecutive frames is a unit time and the time interval ratio is $\lambda$, which can be defined as $t_{0} + t_{1} = 1, t_{1}/t_{0} = \lambda$. Based on the same physical motion model with Eq. \ref{eq:QVI}, the flow can be derived as:
	%\begin{equation}\label{eq:UQVI}
	%\boldsymbol{S_{1t}} = (\frac{\lambda +1}{\lambda}\boldsymbol{S_{12}} - (\lambda + 1)\boldsymbol{S_{01}}) \times t^{2} + (\frac{1}{\lambda} \boldsymbol{S_{12}} + \lambda \boldsymbol{S_{01}}) \times t.
	%\end{equation}
	%This formula enables the interpolation within the video sequence which has uneven time intervals.

	However, Eq.(\ref{eq:QVI}) is based on the equal time interval assumption. This assumption is not applicable to the general situation where the time intervals $ t_{0} $ and $ t_{1} $ may vary accordingly. Here, we define a shutter period as one unit time, and the ratio between $ t_{1} $ and $ t_{0} $ is $ \lambda $, \textit{i.e.}  $t_{0} + t_{1} = 1, \  t_{1}/t_{0} = \lambda$. 
	%According to the derivation of Eq. \ref{eq:QVI}, given three key-points and a fixed time interval, we can determine a unique curvilinear movement, \textit{i.e.} fixed velocity and acceleration. Naturally, if the time intervals become an unknown variable, we will need another key-point to determine a unique movement. So, 
	Different from \cite{xu2019quadratic} which employs three neighboring frames to calculate the quadratic trajectory, we take four consecutive key-states into consideration as shown in Fig.~\ref{fig:exposure}~(b). Naturally, if the time intervals become unknown, four key-states (\textit{i.e.} three flows) are requested to determine a unique quadratic movement. If we assume the acceleration remains constant from frame $ L_{0} $ to  $ L_{3} $, then we can express $\boldsymbol{S_{01}}, \boldsymbol{S_{12}}$ and $ \boldsymbol{S_{23}} $ with velocity and acceleration:
	%\begin{equation}
	\begin{align}
	2\boldsymbol{S_{12}} &= (2\boldsymbol{v_{1}} + \boldsymbol{a}t_{1}) \times t_{1} ,\nonumber \\
	\boldsymbol{S_{01}} + \boldsymbol{S_{23}} &= (2\boldsymbol{v_{1}} + \boldsymbol{a}t_{1}) \times t_{0}.  
	\label{eq:relation}
	\end{align}
	This equation set indicates that vector $ \boldsymbol{S_{12}} $ has a same direction with vector resultant $ \boldsymbol{S_{01}} + \boldsymbol{S_{23}} $. In addition, we can derive the time interval ratio $ \lambda $ as:
	\begin{equation}\label{eq:lambda}
	\lambda = \frac{t_1}{t_0} = \frac{2 \boldsymbol{S_{12}}}{ \boldsymbol{S_{01}} + \boldsymbol{S_{23}}}.
	\end{equation}
	By far, we are able to solve the $ t_{0} $ and $ t_{1} $ under the condition that $t_{0} + t_{1} = 1$. Further deriving the velocity and acceleration of the movement, we can get the expression of trajectory between frame $ L_{1} $ and frame $ L_{2} $:
	\begin{equation}\label{eq:UTI}
	\boldsymbol{S_{1t}} = (\lambda + 1)(\boldsymbol{S_{23}} - \boldsymbol{S_{01}} )/2 \times t^2 + (\lambda \boldsymbol{S_{01}} + (\boldsymbol{S_{01}}+\boldsymbol{S_{23}})/2) \times t, t \in (t_{L_0},t_{L_1}).
	\end{equation}
	
	%As we can see, by introducing another point/frame, we can define the implicit time interval ratio and achieve the interpolation in uncertain time interval sequences. 
	Note that when the time intervals are equal, \textit{i.e.} $ \lambda=1 $, our Eq.(\ref{eq:UTI}) can be degraded to Eq.(\ref{eq:QVI}), \textit{i.e.} the QVI interpolation~\cite{xu2019quadratic} is a special case of our framework.
	
	\subsection{Optical flow refinement}
	As shown in Eq.(\ref{eq:lambda}) (\ref{eq:UTI}), we can obtain flow $ \boldsymbol{S_{1t}} $ using the pixel displacements among four key-states. %Also, we mentioned how to calculate pixel displacements using optical flow. 
	In order to keep the position aligned, $ \boldsymbol{S_{23}} $ should be represented as:
	\begin{equation}\label{eq:s23}
	\boldsymbol{\hat{S}_{23}} = \boldsymbol{f_{1 \rightarrow 3}} - \boldsymbol{f_{1 \rightarrow 2}},
	\end{equation}
	which denotes the movements of pixels in frame $ L_{1} $ from time $ t_{L_2} $ to $ t_{L_3} $. In practical, we estimate all the optical flows using the state-of-the-art PWC-Net \cite{sun2018pwc}. However, directly using flow calculated by Eq.(\ref{eq:s23}) does not work well in our situation, since there may exist serious errors in two aspects: 1) the flow estimation error owing to the long time interval between frame 1 and frame 3, \textit{i.e.} $\boldsymbol{f_{1 \rightarrow 3}}$; 2) the pixel misalignment when we conduct vector subtraction. 
	
	Therefore, we propose a flow refinement network $\mathcal{F_{R}}$ to acquire the refined flow $\boldsymbol{\hat{S}_{23}^{'}}$. Since it is hard to obtain the ground-truth of the target flow $ \boldsymbol{S_{23}} $, we employ the trajectory prior implied in Eq.(\ref{eq:relation}) (\ref{eq:lambda}) as our penalization. Specifically, 
	%So, we manage to refine the flow $ \boldsymbol{S_{23}} $ by utilizing the relation in Eq.(\ref{eq:relation}) (\ref{eq:lambda}) as a trajectory prior. As indicated from Eq.(\ref{eq:relation}) (\ref{eq:lambda}), 
	$ \boldsymbol{\hat{S}_{01}} + \boldsymbol{\hat{S}_{23}} $ and $ \boldsymbol{\hat{S}_{12}} $ should have following two implicit constraints: 1) these two vectors have the same direction; 2) since $\lambda$ is a constant, the ratio of the two vectors should be uniform across the image. With these two priors as constraints, we are able to correct the value of one optical flow when the other two are fixed.
	Note that, although $ \boldsymbol{f_{1\rightarrow0}} $ and $ \boldsymbol{f_{1\rightarrow2}} $ are also the estimation of pixel displacements, they could deliver more accurate motion estimation than $\boldsymbol{\hat{S}_{23}}$. 
	%Therefore, we choose  $ \boldsymbol{\hat{S}_{23}}$ as the refine object. 
	Therefore, the refinement process can be formulated as:
	\begin{equation}
	\boldsymbol{\hat{S}_{23}^{'}} = \mathcal{F_{R}}(\boldsymbol{f_{1\rightarrow0}}, \boldsymbol{f_{1\rightarrow2}}, \boldsymbol{\hat{S}_{23}}).
	\end{equation}
	%where $ \boldsymbol{\hat{S}_{23}^{'}} $ denotes the estimated $ \boldsymbol{S_{23}}$ after refined. Specifically, 
	We use a U-Net \cite{ronneberger2015u} with skip connections to learn the mapping from the original flow to the refined outputs. Aforementioned priors are implicitly encoded into our loss function, where we utilize $ \boldsymbol{f_{0\rightarrow1}}$ and $\boldsymbol{f_{1\rightarrow2}}  $ to constrain the output flow. The loss function is calculated as:
	\begin{equation}
	\mathcal{L}_{r} = |\boldsymbol{\hat{S}_{23}} - (2/\lambda \boldsymbol{f_{1\rightarrow2}} + \boldsymbol{f_{1\rightarrow0}}) |_{1}.
	\end{equation}
	Finally, the refined $ \boldsymbol{\hat{S}_{23}^{'}} $ can be substituted into Eq.(\ref{eq:UTI}) to compute a more accurate $ \boldsymbol{S_{1t}} $.
	
	%\section{Generalized Video Interpolation Scheme}\label{Sec:3}
	%As aforementioned, the flow prediction requires four consecutive key-stateses. In this section, we will discuss how to acquire these key-stateses in a video with unknown exposure settings, and the whole generalized video interpolation scheme.

	\begin{figure}[t!]
		\centering
		\includegraphics[width=\linewidth]{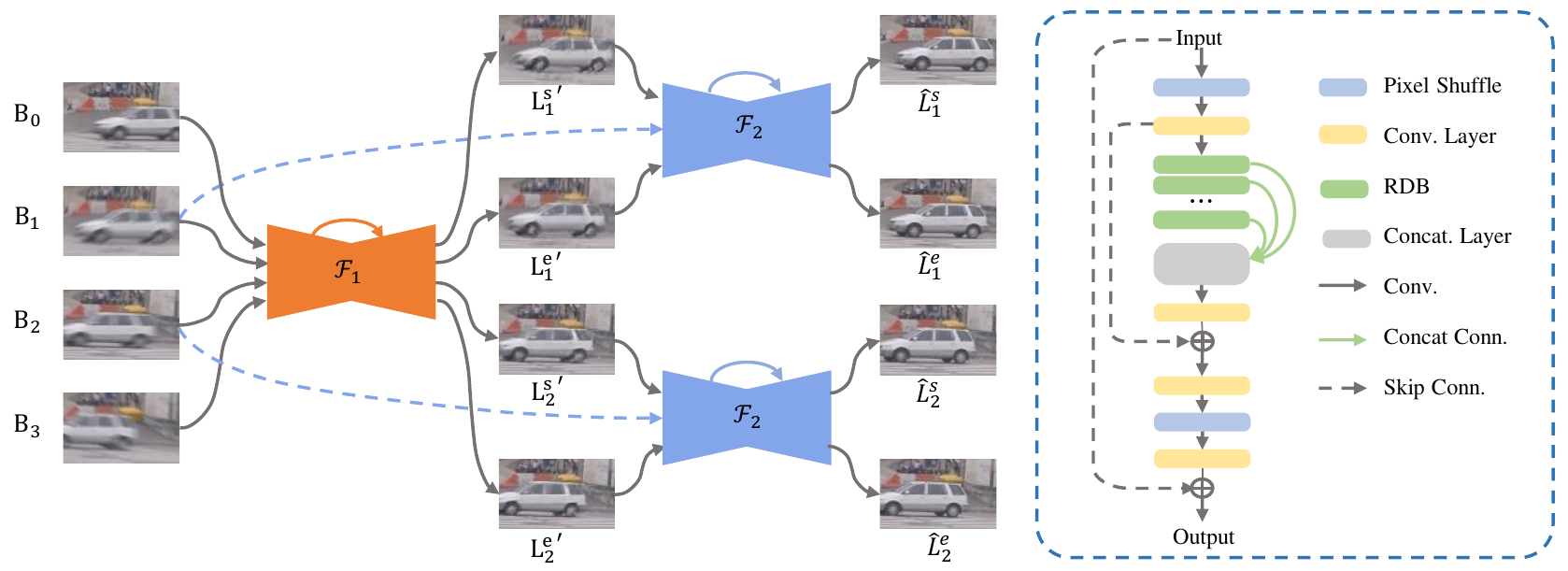}
		\caption{\textbf{Overview of our key-states restoration network.} The left figure shows the inputs and outputs of each sub-network. The right figure shows the backbone structure of $\mathcal{F}_{1}$ and $\mathcal{F}_{2}$.}
		\label{fig:diagram}
	\end{figure}
	
	\subsection{Second-order residual learning for key-states restoration}
	%As former methods \cite{jin2019learning,shen2020blurry} mentioned, jointly deblurring and interpolation/up-convert frames have a better performance than deblurring and interpolation sequentially. Therefore, these methods both choose to use generative model in both deblurring phase and interpolation phase. 
	
	%Actually, the reason for the performance degradation of the cascade scheme is mainly attributed to the two consecutive deblurred key-stateses, \textit{i.e.} middle frames of blurry frames, have a long temporal scope which will cause serious error during optical flow based interpolation. Theoretically, if we shorten the temporal scope of adjacent frames, it will benefit to the optical flow estimation and thus improve the overall interpolation quality. In our proposed model, we manage to shorten this temporal scope by first generating the start and end frames from the blurry frames. There is also a rationale for choosing these two frames as key-stateses for subsequent interpolation operation.
	Our principle for choosing key-states is to ensure that they are unambiguous under different exposure settings.
	%In other words, we aim to find invariant frames regardless of how many intermediate frames are interpolated within the blurry frames or between the blurry frames. 
	For each input frame, we attempt to restore its instant states of the start and end of the exposure, since their physical meaning is consistent in different exposure settings. For sharp images (\textit{i.e.} without motion blur), the start and end states should be the same. For blurry frames, the start and end states define the boundary of the motion blur, which makes them easier to restore. In addition, ours could short the temporal range for subsequent interpolation, which leads to more accurate interpolation results. More discussion can be found in the experiment section.

	%For example, the start and end frame indexes of $ B_{1} $ in Fig.~\ref{fig:exposure}~(a) are 1 and 7, and if the blurry frame is composed of 5 latent frames, the start and end frame indexes will be 1 and 5. 
	%The benefit of choosing these key-stateses are three folds: the first and foremost is that they are unambiguous to different exposure settings. Second, they shorten the temporal scope for subsequent interpolation, which leads to more accurate interpolation results. Third,  the start and end frames define the boundary of the motion blur, which is easier to restore.
	
	As shown in Fig.~\ref{fig:diagram}, we propose the second-order residual learning pipeline to extract the key-states from input frames. Firstly, in order to avoid the temporal ambiguity of the start and end states, four consecutive frames are fed into the network $\mathcal{F}_{1}$. Utilizing the implicit motion direction existed in the input sequence, the network is trained to synthesize residuals to be summed up with input blurry frames, and deliver the start and end states of $B_1$ and $B_2$. This process can be formulated as:
	%we adopt a similar network structure with \cite{jin2019learning,zhang2018residual} which has been commonly used in image deblurring and super-resolution tasks. The main feature extracting part is the cascade of several residual dense blocks (RDB) \cite{zhang2018residual}. 
	%The end of the network will output residuals to be summed up with input blurry frames. This process can be formulated as:
	\begin{equation}\label{first-order}
	(\hat{L}_{i}^{s}, \hat{L}_{i}^{e}) = \mathcal{F}_{1}(B_{seq}) + B_{i}, \ i=1,2,
	\end{equation}
	where $ (\hat{L}_{i}^{s}, \hat{L}_{i}^{e})$ denotes the the estimated instant start and end states respectively, and $ B_{seq} $ denotes the input sequence $\{B_0, \cdots, B_3\}$. 
	
	In the experiments, although the network $\mathcal{F}_1$ achieves reasonable performance, we find it still suffers from some limitations. Firstly, the network gets four inputs to eliminate the temporal ambiguity, which decreases its deblurring capability more or less. Secondly, the fitting ability of residual is relatively poor when modeling a more severe blur. To address these issues, we further improve the deblurring performance by introducing the second-order residual learning. %Motivated by \cite{gao2019dynamic} which validates that fitting the residual of residuals is easier than fitting the residual of original map, we believe that higher-order residual can also be exploited in fitting the output image rather than only the feature map. 
	Specifically, we refer the Eq.(\ref{first-order}) as first order residual, and derive the second-order residual learning as:
	\begin{equation}
	(\hat{L}_{i}^{s}, \hat{L}_{i}^{e}) = \mathcal{F}_{2}(B_{i}, \mathcal{F}_{1}(B_{seq}) + B_{i}) + \mathcal{F}_{1}(B_{seq}) + B_{i}, \       i=1,2.
	\end{equation}
	Here, the network $ \mathcal{F}_{2} $ aims to synthesize higher order residual of the target mapping. Since the temporal order of $ \hat{L}_{i}^{s} $ and $ \hat{L}_{i}^{e} $ has been initially determined by function $ \mathcal{F}_{1} $, $ \mathcal{F}_{2} $ can focus on restoring a pair of key-states. In experiments, this structure could improve the PSNR by around 1.5 dB.

	\section{Experiments}
	In this section, we introduce the datasets we used for training and test, and the training configuration of our models. Then we compare the proposed framework with state-of-the-art methods both quantitative and qualitative. Finally, we carry out an ablation study of our proposed components.
	
	\subsection{Datasets}
	%Since our task need to be evaluated in video clips with different exposure time settings, similar to the approach of synthesis blurry video/frame in \cite{nah2017deep,su2017deep}, 
	To simulate the real-world situations and build datasets for more general video interpolation, we synthesize low-frame-rate videos from the sharp high-frame-rate video sequence. Considering the video acquisition principle we discussed before, we average several consecutive frames taken by a 240fps camera to simulate one frame taken by a low-frame-rate camera. Similar to all the existing blurry image/video datasets, such synthesis is feasible if the relative motion between camera and object is not too large to produce the 'ghosting' artifacts. Meanwhile,
	%In order to conform with the photography law we discussed in Sec. \ref{Sec:Intro}, 
	we discard several consecutive frames to simulate the shutter closed time interval. In this way, we create videos filmed in different exposure settings by altering the number of frames averaged and discarded. Specifically, we denote the number of exposure frames as $ m $, and the number of discarded frames as $ n $, thus $ m+n $ frames constitute a shutter period. We set $ m+n=10 $ to down-sample the original 240 fps video to 24 fps, which is a common FPS setting in our daily life. %Then, we respectively set $ m=5,m=7,m=9 $ to simulate different exposure settings. %Though $ m $ can be any integer from 1 to 10, we choose these three settings for two reasons: (1) A too small $ m $ will lead to slight blur which is not suitable for comparison of deblurring methods. (2) 
	%Here, we set $ m $ as odd number because almost other methods need the temporal center frame as groundtruth, while we do not, so we set $ m $ as odd number for fair comparison. 
	For fair comparisons, we set $ m $ as odd numbers ($m=5, \ 7,\ 9$), since most other methods request the middle frame as ground-truth.
	
	We apply the synthetic rule on both GoPro dataset \cite{nah2017deep} and Adobe240 dataset \cite{su2017deep}, and name these synthetic datasets as ``dataset-m-n''. Finally we get ``GoPro-5-5'', ``GoPro-7-3'', ``GoPro-9-1'' and ``Adobe240-5-5'', ``Adobe240-7-3'', ``Adobe240-9-1'' respectively. 
	%Following official partition, each "GoPro" dataset contains 22 video clips for training and 11 video clips for testing, and each "Adobe240" dataset contains 112 training video clips and 8 testing video clips. %We train and test our model separately on these GoPro and Adobe240 datasets.
	In addition, we also provide datasets ``GoPro-5-3'' and ``GoPro-7-1'' to perform a fair comparison with \cite{shen2020blurry} since it can only upsample the video by multiple of 2. Noted that the other video interpolation datasets such as UCF101 \cite{soomro2012ucf101} and Vimeo-90k \cite{xue2019video} are not applicable for our comparison, since they only provide sharp frame triplets. 
	
	\subsection{Implementation details}
	%For the implementation of network $ \mathcal{F}_1 $ and $ \mathcal{F}_2 $, we adopt same backbone network \cite{jin2019learning} which is a variation of residual dense network \cite{zhang2018residual}. The difference is that we use 20 residual dense blocks in  $ \mathcal{F}_1 $ while 10 in $ \mathcal{F}_2 $. For the implementation of $ \mathcal{F}_{R} $, we adopt U-Net \cite{ronneberger2015u} structure with skip connection. All the detailed network structures are provided in supplementary materials.
	
	To train the key-states restoration network, we first train the network $ \mathcal{F}_1 $ for 200 epochs and jointly train the network $ \mathcal{F}_1 $ and $ \mathcal{F}_2 $ for another 200 epochs. To train the optical flow refinement network, 100 epochs are enough for convergence. We use Adam \cite{kingma2014adam} solver for optimization, with $ \beta_{1}=0.9 $, $\beta_{2}=0.999$ and $\epsilon=10^{-8}$. The learning rate is set initially to $10^{-4}$, and linearly decayed to 0. All weights are initialized using Xavier \cite{glorot2010understanding}, and bias is initialized to 0. In total, we have 34.4 million parameters for training. In test phase, it takes 0.23s and 0.18s to run a single forward for key-states restoration network and interpolation network respectively via a NVIDIA GeForce GTX 1080 Ti graphic card.

	\subsection{Comparison with the state-of-the-art methods}
	\textbf{Comparison methods}. %To perform the generalized video frame interpolation, our framework consists of two important parts, the key-states restoration (deblurring) and following interpolation. 
	We employ two types of interpolation solution as our comparisons.
	%We evaluate the proposed method in two aspects: deblurring performance and overall interpolation performance, which corresponding to the two separate parts of our method: the key frames generation and the uncertain time interval interpolation. 
	%Moreover, we compare our method to two types of interpolation process. 
	The first one is the cascade model, which concatenates a deblurring model with a video frame interpolation model. Specifically, we combine the state-of-the-art image/video deblurring methods Gao \textit{et al.}\cite{gao2019dynamic} and EDVR \cite{wang2019edvr} with the state-of-the-art multi-frame interpolation methods QVI \cite{xu2019quadratic} and Super SloMo \cite{jiang2018super}. We follow the implementation of their official released code in all the experiments.

	The other is the joint model of TNTT and BIN proposed by Jin \textit{et al.} \cite{jin2019learning} and Shen \textit{et al.} \cite{shen2020blurry}, respectively. These methods jointly conduct deblurring and upsampling of frame rate.
	%Actually, the cascade model is a naive implementation to handle the blurry low-frame-rate video and is proven to be sub-optimal compared to the joint model \cite{jin2019learning,shen2020blurry}.
	%For the cascade model,  For joint model, we compare with algorithms TNTT and BIN proposed by Jin \textit{et al.} \cite{jin2019learning} and Shen \textit{et al.} \cite{shen2020blurry}, respectively. 
	Since these two methods are devised for specific exposure setting, we make some workarounds to carry out a more fair and reasonable comparison. Since the original TNTT \cite{jin2019learning} model need to iteratively interpolate the middle frame to fill the vacant indexes, we devise a specific interpolation sequence for each exposure setting, namely TNTT*. 
	%So we revise the output frames to be the start, middle and end frames of the inputs, and re-train the deblurring module. Moreover, the second stage of \cite{jin2019learning} interpolate the middle index between two obtained indexes, namely interpolate the middle frame, yet the interpolation indexes are different between various exposure settings, so we re-order the interpolation process to conform with the specific situation. Both the results of original solution of \cite{jin2019learning} and our modified version (TNTT*) are provided in Table \ref{tb:Gopro results}. 
	In addition, since BIN \cite{shen2020blurry} is devised to up-convert the frame rate by 2 times, which shares the similiar function with our key-states restoration module, we compare their initial results with our first stage outputs. For multi-frame interpolation results, we iteratively interpolate the outputs of BIN and obtain the ``8x frame rate'' results. For this comparison, we prepare the dataset ``5-3'' and ``7-1'' as two different exposure setting of 30 fps video. We re-train and test the BIN model using the mixed datasets ``5-3'' and ``7-1''. Yet, our model is only trained on the mix datasets of ``5-5'', ``7-3'' and ``9-1''. Here, we also test our well-trained model on the ``5-3'' and ``7-1'' settings, experiments in Table~\ref{tb:BIN} shows the great generalization ability of our proposed framework. 
	
	\begin{table}
	\begin{center}
	\caption{Quantitative comparison on the GoPro datasets~\cite{nah2017deep}.}
	\scriptsize
	\renewcommand{\tabcolsep}{3.5pt} %
		\begin{tabular}{ lcccccccccccc }
			\toprule	
			\multirow{3}{*}[-0.58em]{Method} & \multicolumn{6}{c}{Deblurring} & \multicolumn{6}{c}{Interpolation}  \\
			\cmidrule(lr){2-7}
			\cmidrule(lr){8-13}		
			&\multicolumn{2}{c}{GoPro-5-5} & \multicolumn{2}{c}{GoPro-7-3} & 		 			\multicolumn{2}{c}{GoPro-9-1} & \multicolumn{2}{c}{GoPro-5-5} & \multicolumn{2}{c}{GoPro-7-3} & \multicolumn{2}{c}{GoPro-9-1}  \\
			\cmidrule(lr){2-3} \cmidrule(lr){4-5} \cmidrule(lr){6-7}
			\cmidrule(lr){8-9} \cmidrule(lr){10-11} \cmidrule(lr){12-13}
			& PSNR & SSIM & PSNR & SSIM & PSNR & SSIM 
			& PSNR & SSIM & PSNR & SSIM & PSNR & SSIM \\
			\midrule
			EDVR + SloMo & \multirow{2}{*}{31.97} & \multirow{2}{*}{0.9448} & \multirow{2}{*}{29.60} & \multirow{2}{*}{0.9399} & \multirow{2}{*}{28.69}  & \multirow{2}{*}{0.9225} & 27.74 & 0.9010 & 27.09 & 0.9013 & 26.71 & 0.8906 \\
			EDVR + QVI & & & & & & & 28.57 & 0.9152 & 27.42 & 0.9132 & 27.21 & 0.9007  \\
			Gao~\cite{gao2019dynamic} + SloMo & \multirow{2}{*}{\second{32.58}} & \multirow{2}{*}{0.9647} & \multirow{2}{*}{\first{32.64}} & \multirow{2}{*}{0.9674} & \multirow{2}{*}{\second{31.51}} & \multirow{2}{*}{\second{0.9586}} & 28.22 & 0.9086 & 28.31 & 0.9101 & 27.97 & 0.9050 \\
			Gao~\cite{gao2019dynamic} + QVI & & & & & & & \second{29.13} & \second{0.9255} & 29.2 & 0.9263 & 28.5 & 0.9113 \\
			\midrule 
			TNTT \cite{jin2019learning}& 26.78 & 0.8934 & 28.4 & 0.9185 & 30.15  & 0.9383 & 25.29 & 0.8335 & 27.94 & 0.9052 & 30.29 & 0.9398 \\
			\midrule
			TNTT* & 32.49 & \second{0.9660} & 31.45 & 0.9580 & 30.92 & 0.9526 & 28.39 & 0.8660 & \second{30.92} & \second{0.9486} & \second{30.82} & \second{0.9479} \\
			\midrule
			Ours & \first{34.00} & \first{0.9758} & \second{32.63} & \first{0.9674} & \first{31.72} & \first{0.9597} & \first{32.47} & \first{0.9658} & \first{31.95} & \first{0.9628} & \first{30.95} & \first{0.9536}  \\
			\bottomrule
		\end{tabular}
	
	\label{tb:Gopro results}
	\end{center} 
	\vskip -0.4 cm
	\end{table}
	
	\begin{table}
		\begin{center}
			\caption{Quantitative comparison on the Adobe240 datasets~\cite{su2017deep}.}
			\scriptsize
			\renewcommand{\tabcolsep}{3.5pt} %
			\begin{tabular}{ lcccccccccccc }
				\toprule	
				\multirow{3}{*}[-0.58em]{Method} & \multicolumn{6}{c}{Deblurring} & \multicolumn{6}{c}{Interpolation}   \\
				\cmidrule(lr){2-7}
				\cmidrule(lr){8-13}		
				&\multicolumn{2}{c}{Adobe240-5-5} & \multicolumn{2}{c}{Adobe240-7-3} & 		 		\multicolumn{2}{c}{Adobe240-9-1} & \multicolumn{2}{c}{Adobe240-5-5} & \multicolumn{2}{c}{Adobe240-7-3} & \multicolumn{2}{c}{Adobe240-9-1}  \\
				\cmidrule(lr){2-3} \cmidrule(lr){4-5} \cmidrule(lr){6-7}
				\cmidrule(lr){8-9} \cmidrule(lr){10-11} \cmidrule(lr){12-13}
				& PSNR & SSIM & PSNR & SSIM & PSNR & SSIM 
				& PSNR & SSIM & PSNR & SSIM & PSNR & SSIM  \\
				\midrule
				EDVR + SloMo & \multirow{2}{*}{31.97} & \multirow{2}{*}{0.9478} & \multirow{2}{*}{29.96} & \multirow{2}{*}{0.9254} & \multirow{2}{*}{28.49} & \multirow{2}{*}{0.9051} & 28.82 & 0.9204 & 27.85 & 0.9043 & 27.02 & 0.8885  \\
				EDVR + QVI & & & & & & & \second{29.5} & \second{0.9291} & 28.36 & 0.9111 & 27.42 & 0.8937  \\
				Gao~\cite{gao2019dynamic} + SloMo & \multirow{2}{*}{29.39} & \multirow{2}{*}{0.9297} & \multirow{2}{*}{28.98} & \multirow{2}{*}{0.9246} & \multirow{2}{*}{28.45} & \multirow{2}{*}{0.9182} & 27.51 & 0.9057 & 27.35 & 0.9038 & 27.15 & 0.9008 \\
				Gao~\cite{gao2019dynamic} + QVI & & & & & & & 27.95 & 0.9142 & 27.77 & 0.9118 & 27.53  & 0.9080   \\
				\midrule 
				TNTT \cite{jin2019learning}& 28.75 & 0.9277 & 30.85 & 0.9381 & 29.01 & 0.9222 & 26.76 & 0.8831 & 29.10 & 0.9207  & 28.23 & 0.9148  \\
				\midrule
				TNTT* & \second{32.55} & \second{0.9574} & \second{31.76} & \second{0.9529} & \second{30.91} & \second{0.9438} & 28.94 & 0.8836 & \second{31.43} & \second{0.9477} & \second{30.48} & \second{0.9418} \\
				\midrule
				Ours & \first{34.63} & \first{0.9701} & \first{33.06} & \first{0.9617} & \first{32.21} & \first{0.9562} & \first{31.79} & \first{0.9565} & \first{31.52} & \first{0.9529} & \first{30.66} & \first{0.9458} \\
				\bottomrule
			\end{tabular}
			
			\label{tb:Adobe results}
		\end{center} \vskip -0.4 cm	
	\end{table}
	
	\textbf{Blurry video interpolation.} As shown in Table \ref{tb:Gopro results}, Table \ref{tb:Adobe results} and Table \ref{tb:BIN}, both our deblurring and overall interpolation perform favorably against former methods. In addition, several important observations can be made from these results. Firstly, in the deblurring phase, former video deblurring methods encounter great difficulties in maintaining a promising performance in our datasets with different exposure settings. For example, the original TNTT which is trained on ``GoPro-9-1'' performs inferior in generalizing to other test sets. Moreover, even trained on our mixed datasets, the EDVR deteriorates significantly from dataset ``5-5'' to datasets ``7-3'' and ``9-1''. 
	%Also, it is noteworthy that the original TNTT which is trained on "GoPro-9-1" performs inferior in generalizing to other test sets. However, its modified version TNTT* is able to obtain a reasonable performance on all the exposure settings. This comparison also proves the rationality of our proposed unambiguous key-states (start and end frames) generation \first{we need discuss here. if you want point that start end states is important, I don't think it is related to from "GoPro-9-1" to train mix}. %Second, we evaluate the final interpolation results. We can see from the Table~\ref{tb:Gopro results} and Table~\ref{tb:Adobe results}, 
	For the final interpolation results, we can see that cascade models are sub-optimal for the overall performance. Although the deblurring module achieve a high score in PSNR, there is about 3 dB loss in the following interpolation stage. This may be mainly caused by the long temporal scope between two consecutive input frames. Similar conclusion is also obtained in work of \cite{jin2019learning,shen2020blurry}. On the contrary, the joint models usually can achieve a more accurate interpolation results. However, we observe that the interpolation performance of TNTT/TNTT* deteriorates heavily in the exposure setting ``5-5'' (from 32.49 to 28.39 for GoPro dataset). This is mainly because of the iterative synthesis of the middle frame may lead to sub-optimal results in the inter-frame interpolation.
	%\first{Although we have tried our best to output the right indexes order, the result shows that it is sub-optimal to use a single model to fulfill the function of "generating the middle index frame". Because the relative position of the input frames and the targeting interpolation frame is always changing. ``we need discuss''} Different from TNTT/TNTT*, our model use a unified scheme for different exposure settings, while maintains a good result.
	Same conclusion can be obtained from Table \ref{tb:BIN}, the BIN model performs inferior when they attempt to further interpolate the middle frame between former outputs.
	
\begin{figure}[t!]
	\centering
	\includegraphics[width=\linewidth]{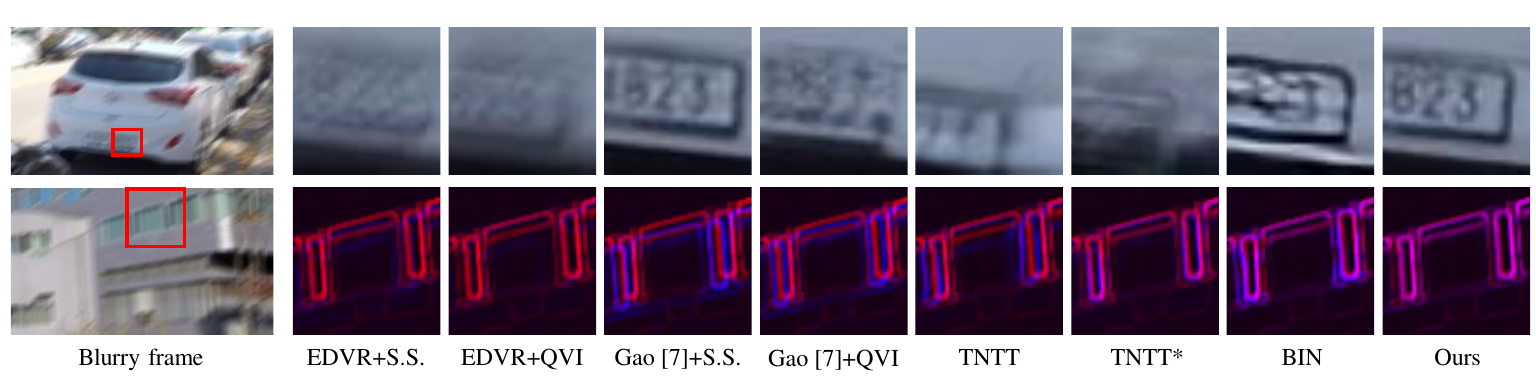} \vskip -0.2 cm
	\caption{\textbf{Visual comparisons on the GoPro dataset.} Each row is an example of the interpolated results from the given blurry frames. S.S. is short for Super slomo \cite{jiang2018super}. The first row shows the original outputs. The second row is the overlap results of the interpolated frame and the ground truth. We extract the contour of each image for a better visual comparison. The red line denotes the contour of ground truth, and the blue line is the outputs of algorithms. The pinker and clearer overlapped image indicates the more accurate interpolation result.}
	\label{fig:comparison}
	\end{figure}

	To intuitively visualize the comparison, we show two typical examples in Fig.~\ref{fig:comparison}. %, and each one corresponds to a factor that influences the interpolation performance. 
	The first row shows that former methods fail in generating a visually clear intermediate frame. This is either because they fail in restoring a sharp frame in deblurring phase, \textit{e.g.} EDVR~\cite{wang2019edvr} and TNTT \cite{jin2019learning}, or the frame becomes blurry when interpolated from adjacent frames, \textit{e.g.} Gao~\cite{gao2019dynamic}+QVI~\cite{xu2019quadratic}, or BIN~\cite{shen2020blurry}. In the second row, we use Sobel operator~\cite{sobel2014history} to extract the contour of interpolated results and overlap it with the contour of ground truth. Red line represents ground-truth contour and blue one means the synthesized outputs. The pinker and clearer overlapped image means a more accurate interpolation result. As we can see,  %there is a obvious misalignment in the results of Gao~\cite{gao2019dynamic}+S.S.~\cite{jiang2018super}/QVI~\cite{xu2019quadratic}, TNTT~\cite{jin2019learning}. Also, we can see a slightly deviation in TNTT* and BIN\cite{shen2020blurry}, while 
	our interpolated frame shows a best overlapped result with ground-truth image.
	
	Moreover, we shot 10 real 30 FPS videos using a telephone camera, and generate the interpolated high-frame-rate video results with our method, as well as TNTT and BIN. Since there is no objective criterion to compare the generation quality, a user study is conducted for a fair comparison. According to more than 1k response collected from Amazon Mechanical Turk, there are 78.4\% of people think our results are better than TNTT's, and 87.6\% of people prefer ours over BIN's results. The real world video interpolation results are provided in our supplementary video.
	%The second row shows that even the interpolated frame is relatively sharp, there is frame misalignment issue caused by inaccurate motion estimation/wrong interpolated index. W
	
	\begin{table}
	\centering
	\caption{Quantitative comparison with BIN \cite{shen2020blurry} on both GoPro and Adobe240 datasets.}
	\footnotesize
	\renewcommand{\tabcolsep}{3.5pt} %
	\begin{tabular}{ lccccccccc }
		\toprule	
		 & \multirow{2}{*}[-0.3em]{Method} & \multicolumn{2}{c}{GoPro-5-3} & \multicolumn{2}{c}{GoPro-7-1}  & \multicolumn{2}{c}{Adobe240-5-3} & \multicolumn{2}{c}{Adobe240-7-1} \\
		\cmidrule(lr){3-4}
		\cmidrule(lr){5-6}
		\cmidrule(lr){7-8}
		\cmidrule(lr){9-10}		
		
		 & & PSNR & SSIM & PSNR & SSIM & PSNR & SSIM 
		& PSNR & SSIM \\
		\midrule
		\multirow{2}{*}{2x frame rate} & BIN & 33.4 & 0.9649 & 32.91 & 0.9675 & 31.35 & 0.9427 & 29.91 & 0.9274  \\
		 & Ours & \first{33.98} & \first{0.9771} & \first{32.96} & \first{0.9707} & \first{33.18} & \first{0.9636} & \first{32.65}  & \first{0.9606}  \\
		\midrule
		\multirow{2}{*}{8x frame rate} & BIN & 30.81 & 0.9553 & 29.14 & 0.9358 & 30.48 & 0.9402 & 29.33 & 9.9244 \\
		& Ours & \first{33.21} & \first{0.9733} & \first{32.3} & \first{0.9667} & \first{32.33} & \first{0.9611} & \first{31.85} & \first{0.9569}  \\
		\bottomrule
	\end{tabular}
	\label{tb:BIN}
	\end{table}

	\textbf{Uncertain time interval interpolation (sharp frames).} To futher validate the effectiveness of our proposed uncertain time interval interpolation algorithm. We compare different interpolation strategies when calculating the essential flow $ \boldsymbol{S_{1t}} $.  To construct the videos with different time interval ratios, we sample the original high-frame-rate GoPro dataset with a different sample interval, \textit{e.g.} to sequentially sample one frame with intervals of 6 frames and 2 frames to achieve the dataset of $ \lambda=7/3 $. We compare our uncertain time interval algorithm (Model UTI) and refined version (Model UTI-refine) with original QVI \cite{xu2019quadratic} model, which is derived under $ \lambda=1 $. We also provide a model GT as the optical flow calculated with ground-truth $ \lambda $. Table \ref{tb:UTI} shows that our UTI and UTI-refine performs favorably to QVI model except the situation when $ \lambda=5/5 $, which is owning to the optical flow estimation error in $ \boldsymbol{S_{23}} $. However, we can see the performance of QVI deteriorates more severe than ours when the value of $ \lambda $ deviates from 1. Also, the results show that our refine network significantly improves the performance.
	
	\begin{table}
		\centering
		\caption{Comparison of different interpolation strategies for uncertain time interval videos.}
		\footnotesize
		\renewcommand{\tabcolsep}{3.5pt} %
		\begin{tabular}{ lcccccc }
			\toprule	
			 \multirow{2}{*}[-0.3em]{Model} & \multicolumn{2}{c}{$ \lambda= 5/5$} & \multicolumn{2}{c}{$ \lambda= 7/3$}  & \multicolumn{2}{c}{$ \lambda= 9/1$}  \\
			\cmidrule(lr){2-3}
			\cmidrule(lr){4-5}
			\cmidrule(lr){6-7}				
			 & PSNR & SSIM & PSNR & SSIM & PSNR & SSIM  \\
			\midrule
			 QVI ($ \lambda=1 $) & 34.68 & 0.9820 & 32.35 & 0.9647 & 29.17 & 0.9239  \\
			\midrule
			 UTI & 33.22 & 0.9691 & 32.65 & 0.9646 & 29.75 & 0.9426 \\
			\midrule
			 UTI-refine & 34.37 & 0.9801 & 33.56 & 0.9739 & 30.57 & 0.9513  \\
			\midrule
			GT & 34.68 & 0.9820 & 34.15 & 0.9787 & 31.34 &  0.9621 \\
			\bottomrule
		\end{tabular}
		\label{tb:UTI}
	\end{table}

	\begin{table}
		\centering
		\caption{Ablation study for key-state restoration and flow refinement.}
		\footnotesize
		\renewcommand{\tabcolsep}{3.5pt} %
		\begin{tabular}{ lccccccc }
			\toprule	
			 & \multirow{2}{*}[-0.3em]{Model} & \multicolumn{2}{c}{GoPro-5-5} & \multicolumn{2}{c}{GoPro-7-3}  & \multicolumn{2}{c}{GoPro-9-1}  \\
			\cmidrule(lr){3-4}
			\cmidrule(lr){5-6}
			\cmidrule(lr){7-8}	
			
			& & PSNR & SSIM & PSNR & SSIM & PSNR & SSIM  \\
			\midrule
			\multirow{4}{*}{Deblurring} & FR & 32.39 & 0.9653 & 31.15 & 0.9547 & 30.38 & 0.9460  \\
			& cascade stage-\uppercase\expandafter{\romannumeral 1\relax} & 32.49 & 0.9644 & 31.41 & 0.9555 & 30.76 & 0.9484 \\
			& Input 2 frames & 33.07 & 0.9700 & 31.83 & 0.9620 & 30.74 & 0.9514 \\
			& Proposed & \textbf{34.00} & \textbf{0.9758} & \textbf{32.63} & \textbf{0.9674} & \textbf{31.72} & \textbf{0.9597} \\
			\midrule
			\multirow{2}{*}{Interpolation} & w/o Refine & 31.82 & 0.9586 & 31.34 & 0.9554 & 30.3 & 0.9434  \\
			& Refine & \textbf{32.47} & \textbf{0.9658} & \textbf{31.95} & \textbf{0.9628} & \textbf{30.95} &  \textbf{0.9536} \\
			\bottomrule
		\end{tabular}
		\label{tb:ablation}
	\end{table}
	
	\subsection{Ablation study}
	To see the effectiveness of our designed modules, we perform the following extensive experiments.
	
	%For the key-state restoration phase, we compare the model using first-order residual as output (Model FR) with the model which use second-order residual (Model SR).
	%As we can see in Table~\ref{tb:ablation}, the second-order residual output can provide over 1.5dB gain in PSNR compared to the first-order residual model. 
	For the key-state restoration phase, we compare the model using different structure/input frames with the proposed model.
	
	As we can see in Table~\ref{tb:ablation}, compared to the first-order residual, the model with second-order residual can increase the PSNR by around 1.5 dB. Also, the model simply cascades another stage-\uppercase\expandafter{\romannumeral 1\relax}'s architecture, \textit{i.e.} without $B_1$, $B_2$ as input, performs inferior to our proposed structure, suggesting the original blurry information is essential for the second-order residual learning. Both the ablation experiments show that our second-order residual is effective in refining the output of the first stage.
	
	For the interpolation phase, we already analyzed the contribution of uncertain time interval interpolation in Table~\ref{tb:UTI}. Here, we evaluate the contribution of the flow refinement module. We fix the key-state restoration network and compare the interpolation outputs of the model with refinement (Model refine) and the model without refinement (Model w/o refine). As shown in Table~\ref{tb:ablation}, the model with refinement outperforms the model without refinement by around 0.6 dB. This improvement indicates that the $ \boldsymbol{\hat{S}_{23}} $ becomes more accurate after refinement.

	\section{Conclusion}
	In this work, we propose a method to tackle the video frame interpolation problem without knowing temporal priors. Taken the relationship of exposure time and shutter period into consideration, we derive a general quadratic interpolation strategy without temporal prior. We also devise a key-states restoration network to extract the temporal unambiguous sharp content from blurry frames. Our proposed method is practical to synthesize a high-frame-rate sharp video from low-frame-rate blurry videos with different exposure settings.
    However, there is still limitation in our work, \textit{e.g.,} our uncertain time interval motion trajectory can only be derived when the acceleration remain constant. Though this assumption can approximate most situations in a short exposure time interval (around 1/20 s), the more challenging movement like variable acceleration motion is existing in the real scenario. We hope to relax this assumption and to have a more accurate trajectory estimation in our future works.%However, there is still limitation in our work, \textit{e.g.} the proposed trajectory priors can only be used to refine one optical flow, while the other flows can also be inaccurate owning to the blurry inputs. We expect the utilization of trajectory priors can be further improved to synthesize a more accurate optical flow in the future work.

	\section*{Broader Impact}

	Video frame interpolation (VFI), which aims to overcome the temporal limitation of camera sensors, is a popular and important technology in a wide range of video processing tasks. For example, it could produce slow-motion videos without professional high-speed cameras, and it could perform the frame rate up-conversion (or video restoration) for archival footage. However, existing VFI researches can mainly apply to videos with pre-defined temporal priors, such as sharp video frames or blurry videos with known exposure settings. It may largely limit their performance in complicated real-world situations. To our best knowledge, the video frame interpolation framework we introduced in this paper made the first attempt to overcome these limitations. %Specifically, in the proposed framework, one well-trained model can be directly adapted to different exposure settings. %It would automatically explore the exposure (and interval) time of the input video and synthesize high-quality slow-motion videos. 
	
	Our proposed technique may potentially benefit a series of real-world applications and users. On the one hand, it could be more practical and convenient for users who want to convert their own videos to slow-motion, since they are not required to figure out the video sources, \textit{i.e.} the complicated parameters of camera sensors. On the other hand, it could reduce the workload of VFI-related applications, \textit{i.e.} it would not need to retrain new models for different exposure settings. 
	
	%Although experiments demonstrate that our framework achieved promising performance in different exposure settings, we would not over-claim the impact of this paper and admit it still faces challenges in the complicated real situations. For example, similar to all other data-driven methods, our model is trained with a subset of data in the real-world. Meanwhile, the acceleration-consistent assumptions between consecutive frames may not fit some real situations. %Moreover, the video frame deblurring performance could also be a bottleneck of final interpolation results. 
	%Overall, this work is only the first attempt to solve the VFI in a more general situation, and it should be carefully employed in complicated real-world situations. Finally, 
	
	Since the video frame interpolation aims at video restoration and up-conversion (\textit{i.e.} the output video shares the same content as the given video), our method may not cause negative ethical impact, if we do not discuss the content of the input video.
	
	\begin{ack}
    This work was supported in part by the Australian Research Council Projects: FL-170100117, DP-180103424, IH-180100002 and IC-190100031.
    \end{ack}
	%\clearpage
	\bibliographystyle{splncs04}
	\bibliography{egbib}
\end{document}